\documentclass[runningheads]{llncs}
\usepackage{graphicx}
\usepackage{tikz}
\usepackage{comment}
\usepackage{amsmath,amssymb} % define this before the line numbering.
\usepackage{color}

\usepackage[accsupp]{axessibility}  % Improves PDF readability for those with disabilities.

\usepackage{bm}
\usepackage{multirow}
\usepackage{pifont}
\usepackage{booktabs}
\usepackage{colortbl}
\usepackage{xspace}
\usepackage[pagebackref,breaklinks,colorlinks]{hyperref}
\usepackage{wrapfig}

\makeatletter
\DeclareRobustCommand\onedot{\futurelet\@let@token\@onedot}
\def\@onedot{\ifx\@let@token.\else.\null\fi\xspace}
 
\def\ie{\emph{i.e}\onedot}

\makeatother

\newcommand{\figref}[1]{Fig.~\ref{#1}}
\newcommand{\tabref}[1]{Tab.~\ref{#1}}
\newcommand{\secref}[1]{\S\ref{#1}}

\newcommand{\equref}[1]{Eq.~(\ref{#1})}

\newcommand{\cmark}{\ding{51}}
\newcommand{\xmark}{\ding{55}}
\DeclareMathOperator*{\argmax}{arg\,max}

\definecolor{tableHeadGray}{gray}{.9}
\newcommand{\PAR}[1]{\noindent{\bf #1}\quad}

\begin{document}
% \linenumbers
\pagestyle{headings}
\mainmatter
\def\ECCVSubNumber{1412}  % Insert your submission number here

\title{Mining Relations among Cross-Frame Affinities for Video Semantic Segmentation} % Replace with your title

\begin{comment}
\titlerunning{ECCV-22 submission ID \ECCVSubNumber} 
\authorrunning{ECCV-22 submission ID \ECCVSubNumber} 
\author{Anonymous ECCV submission}
\institute{Paper ID \ECCVSubNumber}
\end{comment}
%******************

% \begin{comment}
\titlerunning{MRCFA}
\author{Guolei Sun\inst{1} \and Yun Liu\inst{1}\thanks{The corresponding author: yun.liu@vision.ee.ethz.ch} \and Hao Tang\inst{1} \and Ajad Chhatkuli\inst{1} \and \\ Le Zhang\inst{2} \and Luc Van Gool\inst{1,3}\index{Van Gool, Luc}}
\authorrunning{G. Sun et al.}
\institute{Computer Vision Lab, ETH Zurich  \and
School of Information and Communication Engineering, UESTC \and
VISICS, KU Leuven\\
}
% \end{comment}
%******************

\maketitle

\begin{abstract}
The essence of video semantic segmentation (VSS) is how to leverage temporal information for prediction.
Previous efforts are mainly devoted to developing new techniques to calculate the cross-frame affinities such as optical flow and attention.
Instead, this paper contributes from a different angle by  mining relations among cross-frame affinities, upon which better temporal information aggregation could be achieved.
We explore relations among affinities in two aspects: single-scale intrinsic correlations and multi-scale relations.
% By analyzing the properties of cross-frame affinities, we borrow the techniques from traditional feature processing to conduct Single-scale Affinity Refinement (SAR) and Multi-scale Affinity Aggregation (MAA).
Inspired by traditional feature processing, we propose Single-scale Affinity Refinement (SAR) and Multi-scale Affinity Aggregation (MAA).
To make it feasible to execute MAA, we propose a Selective Token Masking (STM) strategy to select a subset of consistent reference tokens for different scales when calculating affinities, which also improves the efficiency of our method.
At last, the cross-frame affinities strengthened by SAR and MAA are adopted for adaptively aggregating temporal information.
Our experiments demonstrate that the proposed method performs favorably against state-of-the-art VSS methods. The code is publicly available at \href {https://github.com/GuoleiSun/VSS-MRCFA}{https://github.com/GuoleiSun/VSS-MRCFA}.
\end{abstract}

\section{Introduction}\label{sec:intro}
Image semantic segmentation aims at classifying each pixel of the input image to one of the predefined class labels, which is one of the most fundamental tasks in visual intelligence.
Deep neural networks have made tremendous progresses in this field \cite{shelhamer2017fully,zhao2017pyramid,chen2018deeplab,huang2019ccnet,zhou2019context,he2019dynamic,he2019adaptive,zhu2019asymmetric,li2020spatial,yuan2020object,jin2021mining,jin2021isnet,ding2018context,ding2019boundary}, benefiting from the availability of large-scale image datasets \cite{cordts2016cityscapes,zhou2019semantic,caesar2018coco,neuhold2017mapillary} for semantic segmentation. However, in real life, we usually confront more complex scenarios in which a series of successive video frames need to be segmented. Thus, it is desirable to explore video semantic segmentation (VSS) by exploiting the temporal information. 

% Deep neural networks have made tremendous progresses in image semantic segmentation, which is one of the most fundamental tasks in vision intelligence. It aims at classifying each pixel of the input image to one of the predefined class labels. %\eg, person, car, and sofa. 
% A number of methods \cite{chen2014semantic,yu2015multi,lin2017refinenet,shelhamer2017fully,zhao2017pyramid,chen2018deeplab,chen2018encoder,zhang2018context,huang2019ccnet,zhou2019context,he2019dynamic,he2019adaptive,zhu2019asymmetric,liu2020learning,zhong2020squeeze,li2020spatial,yuan2020object,jin2021mining,jin2021isnet} have been proposed to promote the advancement of this topic in static images, benefiting from the availability of large-scale image datasets \cite{cordts2016cityscapes,zhou2019semantic,caesar2018coco,neuhold2017mapillary} for semantic segmentation. However, in real-life we usually confront more complex scenarios in which one need to segment a series of successive video frames. Thus, it is desirable to design accurate methods for video semantic segmentation (VSS) by exploiting the temporal information. 

\begin{figure}[!t]
    \centering
    \includegraphics[width=1.0\linewidth]{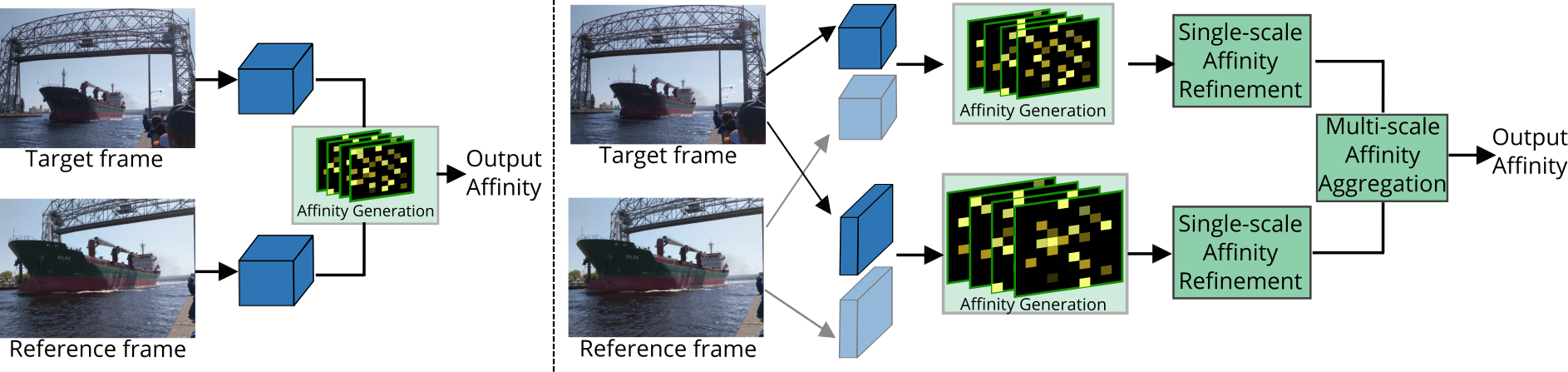}
    % \vspace{-8mm}
    \caption{\textit{Left}: recent VSS methods~\cite{paul2021local,li2021video} for which the affinity is directly forwarded to the next step (feature retrieval). The affinity is shown in a series of 2D maps. \textit{Right}: We propose to mine the relations within the affinities before outputting the affinity, by Single-scale Affinity Refinement (SAR) and Multi-scale Affinity Aggregation (MAA).}
    \label{fig:intro}
    % \vspace{-6mm}
\end{figure}

The core of VSS is how to leverage temporal information. Most of the existing VSS works rely on the optical flow to model the temporal information. Specifically, they first compute the optical flow~\cite{dosovitskiy2015flownet} that is further used to warp the features from neighboring video frames for feature alignment~\cite{zhu2017deep,gadde2017semantic,xu2018dynamic,nilsson2018semantic,jain2019accel,liu2020efficient,lee2021gsvnet}.
Then, the warped features can be simply aggregated.
%In this process, FlowNet \cite{dosovitskiy2015flownet} is usually used for calculating the optical flow by freezing the weights.
Although workable in certain scenarios, those methods are still unsatisfactory because i) the optical flow is error-prone and thus the error could be accumulated; ii) directly warping features may yield inevitable loss on the spatial correlations \cite{li2018low,hu2020temporally}. Hence, other approaches~\cite{paul2021local,li2021video} directly aggregate the temporal information in the feature level using attention techniques, as shown in Fig.~\ref{fig:intro}. Since they are conceptually simple and avoid the problems incurred by optical flow, we follow this way to exploit temporal information. In general, those methods first calculate the attentions/affinities between the target and the references, which are then used to generate the refined features. Though promising, they only consider the single-scale attention. What's more, they do not mine the relations within the affinities.

In this paper, we propose a novel approach MRCFA by Mining Relations among Cross-Frame Affinities for VSS. Specifically, we compute the \textbf{Cross-Frame Affinities (CFA)} between the features of the \textit{target} frame and the \textit{reference} frame.
% Specifically, suppose we want to use the features of \textit{reference} frames to strengthen the features of the \textit{target} frame for improving the target's segmentation, CFA refers to the affinities between the target frame and each reference frame. 
Hence, CFA is expected to have large activation for informative features and small activation for useless features.
When aggregating the CFA-based temporal features, the informative features are highlighted and useless features are suppressed.
As a result, the segmentation of the target frame would be improved by embedding temporal contexts.
With the above analysis, the main focus of this paper is mining relations among CFA to improve the representation capability of CFA. 
Since deep neural networks usually generate multi-scale features and CFA can be calculated at different scales, we can obtain multi-scale CFA accordingly.
Intuitively, the relations among CFA are twofold: single-scale intrinsic correlations and multi-scale relations.

For the \textit{single-scale intrinsic correlations}, each feature token in a reference frame (\ie, reference token) corresponds to a CFA map for the target frame.
Intuitively, we have the observation that the CFA map of each reference token should be locally correlated as the feature map of the target frame is locally correlated, which is also the basis of CNNs.
% ii) At a spatial location, the CFA maps of different reference tokens should be globally correlated as the tokens in the same reference frame are highly correlated.
It is interesting to note that the traditional 2D convolution can be adopted to model such single-scale intrinsic correlations of CFA. Generally, convolution is used for processing features. In contrast, we use convolution to refine the affinities of features for improving the quality of affinities.
We call this step Single-scale Affinity Refinement (SAR).

For the \textit{multi-scale relations}, we propose to exploit the relations among multi-scale CFA maps. The CFA maps generated from high-level features have a small scale and a coarse representation, while the CFA maps generated from low-level features have a large scale and a fine representation.
It is natural to aggregate multi-scale CFA maps using a high-to-low decoder structure so that the resulting CFA would contain both coarse and fine affinities.
Generally, the decoder structure is usually used for fusing multi-scale features. In contrast, we build a decoder to aggregate the multi-scale affinities of features.
We call this step Multi-scale Affinity Aggregation (MAA).

When we revisit the above MAA, one requirement arises: the reference tokens at different scales should have the same number and corresponding semantics; otherwise, it is impossible to connect a decoder. As discussed above, each reference token corresponds to a CFA map for the target frame. Only when two reference tokens have the same semantics, their CFA maps can be merged. For this goal, a simple solution is to downsample reference tokens at different scales into the same size. This also saves the computation due to the reduction of reference tokens.
It inspires us to further reduce the computation by sampling reference tokens.
To this end, we propose a \textbf{Selective Token Masking} strategy to select $S$ most important reference tokens and abandon less important ones.
Then, the relation mining among CFA is executed based on the selected tokens.

In summary, there are three aspects for mining relations among CFA: i) We propose Single-scale Affinity Refinement for refining the affinities among features, based on single-scale intrinsic correlations; 2) We further introduce Multi-scale Affinity Aggregation by using an affinity decoder for aggregating the multi-scale affinities among features; 3) To make it feasible to execute MAA and improve efficiency, we propose Selective Token Masking (STM) to generate a subset of consistent reference tokens for each scale. After strengthened with single-scale and multi-scale relations, the final CFA can be directly used for embedding reference features into the target frame.
Extensive experiments show the superiority of our method over previous VSS methods. Besides, our exploration of affinities among features would provide a new perspective on VSS.

\section{Related Works}
\subsection{Image Semantic Segmentation}
Image semantic segmentation has always been a hot topic in image understanding since it plays an important role in many real applications such as autonomous driving, robotic perception, augmented reality, aerial image analysis, and medical image analysis. In the era of deep learning, various algorithms have been proposed to improve semantic segmentation. Those related works can be divided into two groups: CNN-based methods~\cite{shelhamer2017fully,zhang2019acfnet,chen2020tensor,hsiao2021specialize,chen2018deeplab,yang2018denseaspp,seifi2020attend,sun2020mining,ahn2019weakly,ding2019semantic} and transformer-based methods~\cite{SETR,xie2021segformer}. Among CNN-based methods, FCN~\cite{shelhamer2017fully} is a pioneer work, which adopts fully convolutional networks and pixel-to-pixel classification. 
% Since then, other methods~\cite{chen2014semantic,chen2018deeplab,zhao2017pyramid} have been proposed, with motivations ranging from adopting large receptive fields to designing complex attention modules \cite{huang2019ccnet,zhu2019asymmetric,fu2019dual}. 
Since then, other methods~\cite{chen2014semantic,chen2018deeplab,zhao2017pyramid,huang2019ccnet,zhu2019asymmetric,fu2019dual} have been proposed to increase the receptive fields or representation ability of the network.
Another group of works~\cite{SETR,xie2021segformer} is based on the transformer which is first proposed in natural language processing \cite{vaswani2017attention} and has the ability to capture global context \cite{dosovitskiy2021image}. Though tremendous progress has been achieved in image segmentation, researchers have paid more and more attention to VSS since video streams are a more realistic data modality.  % it is natural to develop methods to segment video streams. 
% Recently, transformers, first proposed in natural language processing, have been widely adopted in various vision tasks~\cite{} due to its ability of capturing global context. 

\subsection{Video Semantic Segmentation}
Video semantic segmentation (VSS), aiming at classifying each pixel in each frame of a video into a predefined category, can be tackled by applying single image semantic segmentation algorithms~\cite{chen2018deeplab,zhao2017pyramid,xie2021segformer,chen2017rethinking,chen2018encoder} on each video frame. Though simple, this approach serves as an important baseline in VSS. One obvious drawback of this method is that the temporal information between consecutive frames is discarded and unexploited. Hence, dedicated VSS approaches~\cite{kundu2016feature,shelhamer2016clockwork,liu2017surveillance,jin2017video,gadde2017semantic,nilsson2018semantic,xiao2018unified,xu2018dynamic,jain2019accel,li2018low,zhu2019improving,liu2020efficient,hu2020temporally,miao2021vspw,paul2021local,paul2020efficient,lee2021gsvnet} are proposed to make use of the temporal dimension to segment videos.

% The core of VSS approaches lies in how the temporal information is used.
Most of the current VSS approaches can be divided into two groups. The first group of approaches focuses on using temporal information to reduce computation. Specifically, LLVS~\cite{li2018low}, Accel~\cite{jain2019accel}, GSVNET~\cite{lee2021gsvnet} and EVS~\cite{paul2020efficient} conserve computation by propagating the features from the key frames to non-key frames. Similarly, DVSNet~\cite{he2019dynamic} divides the current frame into different regions and the regions which do not differ much from previous frames do not traverse the slow segmentation network, but a fast flow network. 
% Though these methods are fast in terms of averaging running time when processing videos, they suffer from the high maximum running time since key frames need to pass through the full segmentation network.
However, due to the fact that they save computation on some frames or regions, their performance is usually inferior to the single frame baseline. 
The second group of methods focuses on exploring temporal information to improve segmentation performance and prediction consistency across frames. 
%Specifically, NetWarp~\cite{xiao2018unified} wraps the features of the reference frames and combines the wrapped representations with the features of the target frame for segmenting the target. 
Specifically, NetWarp~\cite{xiao2018unified} wraps the features of the reference frames for temporal aggregation.
TDNet~\cite{hu2020temporally} aggregates the features of sequential frames with an attention propagation module. 
%TCB~\cite{miao2021vspw} proposes aggregate the contexts in the temporal dimension by a temporal context blending network. 
ETC~\cite{liu2020efficient} uses motion information to impose temporal consistency among predictions between sequential frames. STT~\cite{li2021video},  LMANet~\cite{paul2021local} and CFFM~\cite{sun2022coarse} exploit the features from reference frames to help segment the target frame by the attention mechanism. 
Despite the promising results, those methods do not consider correlation mining among cross-frame affinities. This paper provides a new perspective on VSS by mining the relations among affinities.

% 1. video semantic segmentation can be simply tackled by single image semantic segmentation models. baseline
% 2. dedicated video semantic segmentation can be divided by their emphasis: some methods focus on improving efficiency; some methods focus on modelling the temporal relationships, thus improving performance
% LLVS (Low-Latency Video Semantic Segmentation): full pass on key frames and save computation on non-key frames

\begin{figure*}[!t]
    \centering
    \includegraphics[width=1.0\linewidth]{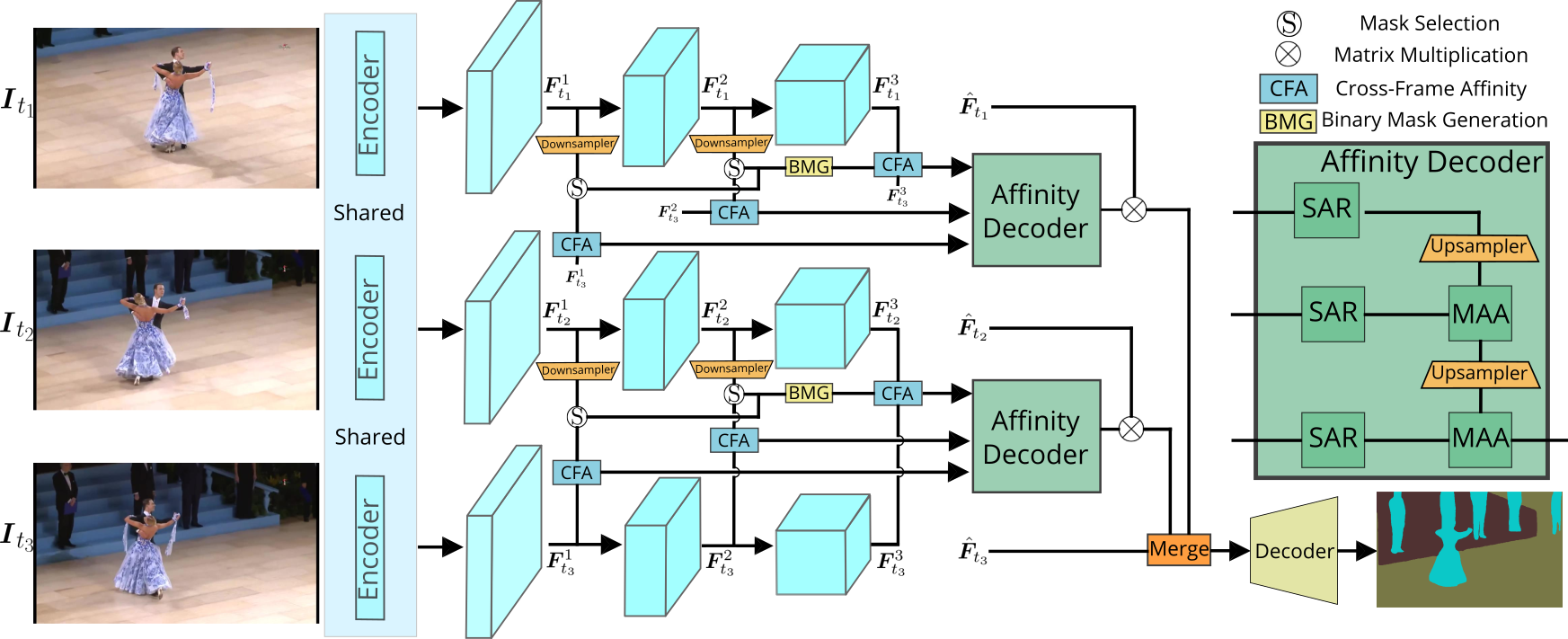}
    % \vspace{-7mm}
    \caption{{{Network overview of MRCFA.}} Our method is illustrated when the clip contains three frames ($T=3$). The first two frames are reference frames while the last one is the target frame. All frames first go through the encoder to extract the multi-scale features ($L=3$) from the intermediate layers. For each reference frame, we compute the Cross-Frame Affinities (CFA) across different scales of features. To save computation, Selective Token Masking is proposed. Then, the multi-scale affinities are input to an affinity decoder to learn a unified and informative affinity, through the Single-scale Affinity Refinement (SAR) module and Multi-scale Affinity Aggregation (MAA). The new representation of the target frame using the reference is obtained by exploiting the refined affinity to retrieve the corresponding reference features. Finally, all the new representations of the target are merged to segment the target. \textit{Best viewed in color.}}
    \label{fig:framework}
\end{figure*}

\section{Methodology}
In this section, we target VSS and present a novel approach MRCFA through Mining Relations among Cross-Frame Affinities. The main idea of MRCFA is to mine the relations among multi-scale affinities computed from multi-scale intermediate features between the target frame and the reference frames, as illustrated in \figref{fig:framework}. We first provide the preliminaries in \secref{sec:method_preliminary}. Next, we introduce Single-scale Affinity Refinement (SAR) which independently refines each single-scale affinity in \secref{sec:method_sar}. After that, Multi-scare Affinities Aggregation (MAA) which merges affinities across various scales is presented in \secref{sec:method_maa}. Finally, we explain the Selective Token Masking mechanism (\secref{sec:method_stm}) to reduce the computation.

% 1. difference between the proposed method and the baseline of feature pyramid. 
% a: the proposed method focuses on generating accurate attention maps between the target frame and the reference frames, while the baseline focuses on using merging multi-scale features with diverse receptive fields to generate better features through pyramid network. 
% b: For the proposed method, the smooth is done in a pixel-to-pixel level (pixels in the target frame to pixels in the reference frames) while the feature smooth for the baseline is only performed for the pixels in target frame. That's to say, the smoothing for the proposed method is more fine-grained than the baseline.

\subsection{Preliminaries}\label{sec:method_preliminary}
Given a video clip $\{\bm{I}_{t_i} \in \mathbb{R}^{H \times W \times 3} \}_{i=1}^{T}$ containing $T$ video frames and corresponding ground-truth masks $\{\bm{M}_{t_i} \in \mathbb{R}^{H \times W} \}_{i=1}^{T}$, our objective is to learn a VSS model. Without loss of generalizability, we focus on segmenting the last frame $\bm{I}_{t_{T}}$, which is referred as the target frame. All the previous frames $\{\bm{I}_{t_i}\}_{i=1}^{T-1}$ are referred as the reference frames. Each frame $\bm{I}_{t_i}$ is first input into an encoder to extract intermediate features $\{\bm{F}_{t_{i}}^{l} \in \mathbb{R}^{H_l W_l \times C_l}\}^{L}_{l=1}$ in various scales from $L$ intermediate layers of the deep encoder, where $H_l$, $W_l$, $C_l$ correspond to the height, width, number of channels of the feature map, respectively. 
% The encoder also outputs an informative feature $\hat{\bm{F}}_{t_{i}} \in \mathbb{R}^{\hat{C} \times \hat{H} \times \hat{W}}$ which can be directly used to predict the segmentation map for each frame.
For simplicity, multi-scale features $\{\bm{F}_{t_{i}}^{l}\}^{L}_{l=1}$ are in the order that shallow features are followed by deep features. 
% $\bm{F}_{t_{i}}^{1}$ and $\bm{F}_{t_{i}}^{L}$ correspond to the shallowest and deepest features, respectively.
We have $H_{l_1}\geq H_{l_2}$ and $W_{l_1}\geq W_{l_2}$, if $l_1<l_2$.
% In our implementation, we use the same backbone as SegFormer~\cite{xie2021segformer} and $\bm{F}_{t_{i}}^{1}, \bm{F}_{t_{i}}^{2}, \bm{F}_{t_{i}}^{3}$ correspond the features output in the last three stages.
In this paper, we aim to exploit the contextual information in the reference frames to refine the features of the target frame and thus improve the target's segmentation. 
Instead of simply modeling the affinities among frames for feature aggregation, we devote our efforts to mine relations among cross-frame affinities.

\subsection{Single-scale Affinity Refinement} \label{sec:method_sar}
% \PAR{Affinity Generation}
We start with introducing the process of generating multi-scale affinities between the target frame and each reference frame.
We first map the features $\{\bm{F}_{t_{T}}^{l}\}^{L}_{l=1}$ of the target frames into the queries $\{\bm{Q}^{l}\}^{L}_{l=1}$ by a linear layer, as:
\begin{equation}\label{Eq:gene_q}
\footnotesize
    \bm{Q}^l=f(\bm{F}_{t_{T}}^{l}; \bm{W}^l_{query}),
\end{equation}
where $\bm{W}^l_{query} \in \mathbb{R}^{C_l \times C_l}$ is the weight matrix of the linear layer $f$ and $\bm{Q}^l \in \mathbb{R}^{H_l W_l \times C_l}$. Similarly, the multi-scale features $\{\bm{F}_{t_{i}}^{l}\}^{L}_{l=1}$ of the reference frame ($i \in [1, T-1]$) are also processed to generate the keys $\{\bm{K}^{l}_{t_{i}}\}^{L}_{l=1}$, as follows:
\begin{equation}\label{Eq:gene_k}
\footnotesize
    \bm{K}^{l}_{t_{i}} = f(\bm{F}_{t_{i}}^{l}; \bm{W}^l_{key}),
\end{equation}
where $\bm{W}^l_{key} \in \mathbb{R}^{C_l\times C_l}$ is the corresponding weight matrix and $\bm{K}^{l}_{t_{i}} \in \mathbb{R}^{H_l W_l \times C_l}$. After obtaining the queries and the keys, we are ready to generate the affinities between the target frame $\bm{I}_{t_{T}}$ and each reference frame $\bm{I}_{t_i}$ ($i \in [1, T-1]$) across all scales.
% To simplify the explanation, we focus our introduction on computing the affinities between the target frame $\bm{I}_{t_{k}}$ and one reference frame $\bm{I}_{t_i}$, where $i \in [1,k-1]$. 
% The affinities for other reference frames can be computed similarly.
%
Then, \textbf{Cross-Frame Affinities (CFA)} are computed as:
\begin{equation}\label{Eq:gene_affinity}
\footnotesize
    \bm{A}^{l}_{t_{i}} = \bm{Q}^{l} \times {\bm{K}^{l\top}_{t_{i}}},
\end{equation}
where we have $\bm{A}^{l}_{t_{i}} \in \mathbb{R}^{H_l W_l \times H_l W_l}$, $l \in [1, L]$ and $i \in [1, T-1]$.
It means that, at each scale, the target frame has an affinity map with each reference frame.
% We will show how to mine the relations among these affinity maps $\bm{A}^{l}_{t_{i}}$ for better aggregating cross-frame contextual information.

Based on the affinities $\{{\bm{A}}^{l}_{t_{i}}\}_{l=1}^{L}$, our affinity decoder is designed to mine the correlations between them to learn a better affinity between the target and the reference frame. As shown in \figref{fig:framework}, it is comprised of two modules: Single-scale Affinity Refinement (SAR) and Multi-scale Affinity Aggregation (MAA). Please refer to \secref{sec:intro} for our motivations. In order to reduce computation and prepare the affinities for MAA module which requires the same number and corresponding semantics (see \secref{sec:intro}), our affinity decoder operates on  $\{\tilde{\bm{A}}^{l}_{t_{i}} \in \mathbb{R}^{H_l W_l \times S}\}_{l=1}^{L}$, rather than $\{{\bm{A}}^{l}_{t_{i}} \in \mathbb{R}^{H_l W_l \times H_l W_l}\}_{l=1}^{L}$. The affinities $\tilde{\bm{A}}^{l}_{t_{i}}$ is a downsampled version of ${\bm{A}}^{l}_{t_{i}}$ along the second dimension, which will be explained in \secref{sec:method_stm}.

% After generating multi-scale affinities $\{\{\tilde{\bm{A}}^{l}_{t_{i}}\}_{l=1}^{L}\}_{i=1}^{T-1}$ between the target frame and all the reference frames, we introduce the affinity decoder to exploit the correlations between the multi-scale affinities to learn better affinities. As shown in \figref{fig:framework}, our affinity decoder is comprised of two modules: Single-scale Affinity Refinement (SAR) and Multi-scale Affinity Aggregation (MAA). To reduce computation and prepare the affinities for MAA module where the same number and corresponding semantics are needed, we 

% As above, we also focus our discussion on processing the affinities $\{\tilde{\bm{A}}^{l}_{t_{i}}\}_{l=1}^{L}$ for one reference frame $\bm{I}_{t_i}$ ($i \in [1, T-1]$). 

\PAR{Single-scale Affinity Refinement (SAR).} 
For the affinity matrix $\tilde{\bm{A}}^{l}_{t_{i}}$, each of its elements corresponds to a similarity between a token in the query and a token in the key. We reshape $\tilde{\bm{A}}^{l}_{t_{i}}$ from $\mathbb{R}^{H_l W_l \times S}$ to $\mathbb{R}^{H_l \times W_l \times S}$. 
In order to learn the correlation within the single-scale affinity $\tilde{\bm{A}}^{l}_{t_{i}} \in \mathbb{R}^{H_l \times W_l \times S}$, a straightforward way is to exploit 3D convolution. However, this approach suffers from two weaknesses. First, it requires a large amount of computational cost. Second, not all the activations within the 3D window are meaningful. Considering a 3D convolution with a kernel $\mathcal{K} \in \mathbb{R}^{k\times k \times k}$, the normal 3D convolution at the location $x=(x_1, x_2, x_3)$ is formulated as:
\begin{align}\label{Eq:affi_3d}
\footnotesize
% \vspace{-5mm}
\begin{split}
    % &\bm{M}_{t_{i}}={\rm sum}_1(\tilde{\bm{A}}^{L}_{t_{i}}),\\
    (\tilde{\bm{A}}^{l}_{t_{i}} * \mathcal{K})_{x}=\sum_{(o_1,o_2,o_3) \in \mathcal{N}(x)} \tilde{\bm{A}}^{l}_{t_{i}}(o_1,o_2,o_3)\mathcal{K}(o_1-x_1,o_2-x_2,o_3-x_3),
\end{split}
% \vspace{-8mm}
\end{align}
where $\mathcal{N}(x)$ is the set of locations in the 3D window ($k\times k \times k$) centered at $x$, and $|\mathcal{N}(x)| = k^3$. As seen in \equref{Eq:affi_3d}, all the neighbors along three dimensions are used to conduct the 3D convolution. However, the last dimension of $\tilde{\bm{A}}^{l}_{t_{i}}$ is the sparse selection in the key (\secref{sec:method_stm}) and thus does not contain spatial information. Including the neighbors along the last dimension could introduce noise and bring more complexity. 
Thus, we propose to refine the affinities across the first two dimension. For affinity $\tilde{\bm{A}}^{l}_{t_{i}}$ of each scale, we first permute it to $\mathbb{R}^{S\times H_l \times W_l}$ and then use 2D convolutions to learn the relations within the affinity. The refined affinity is denoted as $\bar{\bm{A}}^{l}_{t_{i}} \in \mathbb{R}^{S\times H_l \times W_l}$. 
This process can be formulated as:
\begin{equation}
\footnotesize
\begin{aligned}
&\tilde{\bm{A}}^{l}_{t_{i}} \in \mathbb{R}^{H_l \times W_l \times S} \to \tilde{\bm{A}}^{l}_{t_{i}} \in \mathbb{R}^{S\times H_l \times W_l}, \\
% \bar{\bm{A}}^{l}_{t_{i}} = g(g(\tilde{\bm{A}}^{l}_{t_{i}}; (3,3); (1,1)); (3,3); (1,1)), 
&\bar{\bm{A}}^{l}_{t_{i}} = G(\tilde{\bm{A}}^{l}_{t_{i}}), 
\end{aligned}
\end{equation}
where $G$ represents a few connvolutional layers.
% where $g(\cdot;(k_h,k_w);(s_h,s_w))$ represents a convolutional layer with the kernel size $(k_h,k_w)$ and the stride $(s_h,s_w)$. 
Due to the use of 2D convolution and the token reduction mentioned in \secref{sec:method_stm}, the refinement of affinities is fast. After refining affinity for each scale, we collect the refined affinities $\{\bar{\bm{A}}^{l}_{t_{i}}\}^{L}_{l=1}$ for all scales. Next, we present Multi-scale Affinity Aggregation (MAA) module.

\subsection{Multi-scale Affinity Aggregation} \label{sec:method_maa}
\PAR{Multi-scale Affinity Aggregation (MAA).} 
 The affinity from the deep features contains more semantic but more coarse information, while the affinity from the shallow features contains more fine-grained but less semantic information. Thus, we propose a \textbf{Multi-scale Affinity Aggregation} module to aggregate the information from small-scale affinities to large-scale affinities, as:
% \begin{equation}\label{Eq:main_tranformers}
% \begin{aligned}
%     &\bm{B}^{1}_{t_{i}}=\bar{\bm{A}}^{1}_{t_{i}}, \\
%     &\bm{B}^{l}_{t_{i}}=g(\Gamma(\bm{B}^{l-1}_{t_{i}})+\bar{\bm{A}}^{l}_{t_{i}}; (3,3); (1,1)),  ~~~l=[2,L], \\
%     % &T_{i}={\rm MLP}(T_i^{'})+T_i^{'}, ~~~i=1,...,l.\\
% \end{aligned}
% \end{equation}
\begin{equation}\label{Eq:main_tranformers}
\footnotesize
\begin{aligned}
    &\bm{B}^{L}_{t_{i}}=\bar{\bm{A}}^{L}_{t_{i}}, \\
    &\bm{B}^{l}_{t_{i}}=G(\Gamma(\bm{B}^{l+1}_{t_{i}})+\bar{\bm{A}}^{l}_{t_{i}}),  ~~~l=L-1,...,1, \\
    % &T_{i}={\rm MLP}(T_i^{'})+T_i^{'}, ~~~i=1,...,l.\\
\end{aligned}
\end{equation}
where $\Gamma$ denotes upsampling operation to match the spatial size when necessary. 
By \equref{Eq:main_tranformers}, we generate the final refined affinity $\bm{B}^{1}_{t_{i}}$ between the target frame $\bm{I}_{t_{T}}$ and each reference frame $\bm{I}_{t_{i}}$ ($i \in [1, L-1]$).

\PAR{Feature Retrieval.} 
For single-frame semantic segmentation, SegFormer \cite{xie2021segformer} generates the final feature $\hat{\bm{F}}_{t_{i}} \in \mathbb{R}^{\hat{H} \hat{W} \times \hat{C}}$ by merging multiple intermediate features. The final features are informative and directly used to predict the segmentation mask~\cite{xie2021segformer}. Using the refined affinity $\bm{B}^{1}_{t_{i}}$ and the informative features $\hat{\bm{F}}_{t_{i}}$, we compute the new refined feature representations for the target frame. 
Specifically, the feature $\hat{\bm{F}}_{t_{i}}$ is first downsampled to the size of $\mathbb{R}^{H_L W_L \times \hat{C}}$. To correspond the refined affinity and the informative feature, we sample feature $\hat{\bm{F}}_{t_{i}}$ using the token selection mask $\tilde{\bm{M}}_{t_{i}}$ (\secref{sec:method_stm}) and obtain $\tilde{\bm{F}}_{t_{i}} \in \mathbb{R}^{S \times \hat{C}}$. The new feature representation for the target frame using the reference is obtained as:
\begin{equation}\label{Eq:feat_retri}
\footnotesize
\begin{aligned}
    % &\bm{M}_{t_{i}}={\rm sum}_1(\tilde{\bm{A}}^{L}_{t_{i}}),\\
    \bm{B}^{1}_{t_{i}} \in \mathbb{R}^{S\times H_1 \times W_1} \to \bm{B}^{1}_{t_{i}} \in \mathbb{R}^{H_1 W_1 \times S}, \qquad\quad
    \bm{O}_{t_i} = \bm{B}^{1}_{t_{i}} \times {\tilde{\bm{F}}_{t_{i}}}.
\end{aligned}
\end{equation}
Intuitively, this step is to retrieve the informative features from the reference frame to the target frame using affinity. Computing \equref{Eq:feat_retri} for all reference frames, we obtain the new representations of the target frame as $\{\bm{O}_{t_i}\}^{T-1}_{i=0}$. 

The final feature used to segment the target frame is merged from $\{\bm{O}_{t_i}\}^{T-1}_{i=0}$ and $\hat{\bm{F}}_{t_{L}}$ as follows:
\begin{align}
\footnotesize
\begin{split}
    % &\bm{M}_{t_{i}}={\rm sum}_1(\tilde{\bm{A}}^{L}_{t_{i}}),\\
    \bm{O}_{t_L}=\frac{1}{T-1}\Gamma(\sum_{i=1}^{T-1}\bm{O_{t_i}})+\hat{\bm{F}}_{t_{L}}.
\end{split}
\end{align}
Finally, a simple MLP decoder projects $\bm{O}_{t_L}$ to the segmentation logits, and typical cross-entropy loss is used for training. In the test period, when segmenting the target frame $I_{t_{T}}$, the encoder only needs to generate the features for the current target while the reference frames are already processed in previous steps and the corresponding features can be directly used.

\subsection{Selective Token Masking} \label{sec:method_stm}
%
% As mentioned previously, our motivation is to mine the correlations within $\{\bm{A}^{l}_{t_{i}}\}^{L}_{l=1}$ to learn a better affinity between the target frame and each reference frame. 
As discussed in \secref{sec:intro}, there should be the same number of reference tokens with corresponding semantics across scales.
Besides, computing cross-frame affinities requires a lot of computation.
Thus, our affinity decoder does not process $\{{\bm{A}}^{l}_{t_{i}} \in \mathbb{R}^{H_l W_l \times H_l W_l}\}_{l=1}^{L}$, but rather its downsampled version $\{\tilde{\bm{A}}^{l}_{t_{i}} \in \mathbb{R}^{H_l W_l \times S}\}_{l=1}^{L}$. Here, we explain how to generate $\{\tilde{\bm{A}}^{l}_{t_{i}}\}^{L}_{l=1}$, by reducing the number of tokens in the multi-scale keys $\{\bm{K}^{l}_{t_{i}}\}^{L}_{l=1}$ before computing \equref{Eq:gene_affinity}.

We exploit convolutional layers to downsample the multi-scale keys to the spatial size of $H_L \times W_L$. Specifically, for the key $\bm{K}^{l}_{t_{i}}$ ($l \in [1, L-1]$), we process it by a convolutional layer with both kernel and stride size of $(\frac{H_l}{H_L}, \frac{W_l}{W_L})$. As a result, we obtain new keys $\hat{\bm{K}}^{l}_{t_{i}}$ with smaller spatial size, which is given by
\begin{equation}\label{Eq:tr_down}
\footnotesize
\begin{aligned}
    \bm{K}^{l}_{t_{i}} \in \mathbb{R}^{H_l W_l \times C_l} \to \bm{K}^{l}_{t_{i}} \in \mathbb{R}^{C_l \times H_l \times W_l}, \\
    \hat{\bm{K}}^{l}_{t_{i}} = g(\bm{K}^{l}_{t_{i}}; (\frac{H_l}{H_L},\frac{W_l}{W_L}); (\frac{H_l}{H_L},\frac{W_l}{W_L})), \\
    \hat{\bm{K}}^{l}_{t_{i}} \in \mathbb{R}^{C_l \times H_L \times W_L} \to \hat{\bm{K}}^{l}_{t_{i}} \in \mathbb{R}^{H_L W_L \times C_l}.
\end{aligned}
\end{equation}
% where $l \in [1, L-1]$, $\hat{\bm{K}}^{l}_{t_{i}} \in \mathbb{R}^{C_l, H_L, W_L}$, and ${\rm conv}$ represents a convolutional layer.
% where $g(\cdot;(k_h,k_w);(s_h,s_w))$ represents a convolutional layer with the kernel size $(k_h,k_w)$ and the stride $(s_h,s_w)$.
where $g(\cdot;(k_h,k_w);(s_h,s_w))$ represents a convolutional layer with the kernel size $(k_h,k_w)$ and the stride $(s_h,s_w)$.
After this step, we obtain the downsampled keys $\{\hat{\bm{K}}^{l}_{t_{i}}\}^{L-1}_{l=1}$, where $\hat{\bm{K}}^{l}_{t_{i}} \in \mathbb{R}^{H_L W_L \times C_l}$, $l \in [1, L-1]$ and $i \in [1, T-1]$.

To further reduce the number of tokens in $\{\hat{\bm{K}}^{l}_{t_{i}}\}^{L-1}_{l=1}$, we propose to select important tokens and discard less important ones. The idea is to first compute the affinity for the deepest query/key pair ($\bm{Q}^{L}$ and $\bm{K}^{L}_{t_{i}}$), then generate a binary mask of important token locations, and finally select tokens in keys using the mask. The process of \textbf{Binary Mask Generation (BMG)} is in the following. The affinity between the deepest query and key is given by $\bm{A}^{L}_{t_{i}} \in \mathbb{R}^{H_L W_L \times H_L W_L}$, following \equref{Eq:gene_affinity}. Next, we choose the top-$n$ maximum elements across each column of $\bm{A}^{L}_{t_{i}}$, given by
\begin{align}\label{Eq:tr_argmax}
\footnotesize
\begin{split}
    \hat{\bm{A}}^{L}_{t_{i}}[:, j] = \argmax_{n}(\bm{A}^{L}_{t_{i}}[:, j]), \qquad j \in [1,H_L W_L],
\end{split}
\end{align}
where $\argmax_{n}$ means to take the top-$n$ elements, and $\hat{\bm{A}}^{L}_{t_{i}} \in \mathbb{R}^{n \times H_L W_L}$. Then, we sum over the top-$n$ elements and generate a token importance map $\bm{M}_{t_{i}}$ as
\begin{align}\label{Eq:tr_sum}
\footnotesize
\footnotesize
\begin{split}
    \bm{M}_{t_{i}} = \sum_{j=1}^{n}(\hat{\bm{A}}^{L}_{t_{i}}[j, :]),
    % &\bm{M}_{t_{i}} \in \mathbb{R}^{H_L W_L} \xrightarrow{\rm reshape} \bm{M}_{t_{i}} \in \mathbb{R}^{H_L \times W_L},
\end{split}
\end{align}
in which we have $\bm{M}_{t_{i}} \in \mathbb{R}^{H_L W_L}$. We recover the spatial size of $\bm{M}_{t_{i}}$ by reshaping it to $\mathbb{R}^{H_L \times W_L}$. The token importance map $\bm{M}_{t_{i}}$ shows the importance level of every location in the key feature map. Since $\bm{M}_{t_{i}}$ is derived from the deepest/highest level of features, the token importance information it contains is semantic-oriented and can be shared in other shallow levels. We use it to sample the tokens in $\{\hat{\bm{K}}^{l}_{t_{i}}\}^{L-1}_{l=1}$.
Specifically, we sample $p$ percent of the locations with the top-$p$ highest importance scores in $\bm{M}_{t_{i}}$, where $p$ is referred as the token selection ratio. The binary token selection mask with $p$ percent of the locations highlighted is denoted as $\tilde{\bm{M}}_{t_{i}}$. 
The location with the value 1 in $\tilde{\bm{M}}_{t_{i}}$ means the token importance is within the top-$p$ percent and the corresponding token will be selected. The location with the value 0 in $\tilde{\bm{M}}_{t_{i}}$ means the token in that location is less important and will thus be discarded. The total number of locations with the value 1 in $\tilde{\bm{M}}_{t_{i}}$ is denoted by $S=p H_L W_L$.

Using mask $\tilde{\bm{M}}_{t_{i}}$, we select $p$ percent of tokens in $\{\hat{\bm{K}}^{l}_{t_{i}}\}^{L-1}_{l=1}$. The keys after selection are denoted as $\{\tilde{\bm{K}}^{l}_{t_{i}} \in \mathbb{R}^{S \times C_l}\}^{L-1}_{l=1}$. With $\bm{Q}^{l}$ and $\tilde{\bm{K}}^{l}_{t_{i}}$, we compute the affinities $\{\tilde{\bm{A}}^{l}_{t_{i}} \in \mathbb{R}^{H_l W_l \times S}\}^{L-1}_{l=1}$ using \equref{Eq:gene_affinity}. For ${\bm{A}}^{L}_{t_{i}}$, we also conduct sampling using $\tilde{\bm{M}}_{t_{i}}$ and obtain $\tilde{\bm{A}}^{L}_{t_{i}} \in \mathbb{R}^{H_L W_L \times S}$. Merging the affinities from all $L$ scales gives final affinities of $\{\tilde{\bm{A}}^{l}_{t_{i}} \in \mathbb{R}^{H_l W_l \times S}\}^{L}_{l=1}$. 
% To prepare the affinities for the correlation learning which will be introduced in the following, we reshape the affinities $\tilde{\bm{A}}^{l}_{t_{i}}$ as: $ \mathbb{R}^{H_l W_l \times S} \xrightarrow{\rm } \mathbb{R}^{H_l \times W_l \times S}$.
After computing the affinities for all reference frames, we have the downsampled affinities $\{\{\tilde{\bm{A}}^{l}_{t_{i}}\}^{L}_{l=1}\}^{T-1}_{i=1}$. 

% \tilde{\bm{A}}^{l}_{t_{i}} \in \mathbb{R}^{H_l \times W_l \times S} \to \tilde{\bm{A}}^{l}_{t_{i}} \in \mathbb{R}^{S\times H_l \times W_l}, \\
% \bar{\bm{A}}^{l}_{t_{i}} = g(g(\tilde{\bm{A}}^{l}_{t_{i}}; (3,3); (1,1)); (3,3); (1,1)).
% \end{aligned}
% \end{equation}
% Due to the use of 2D convolution and the token reduction mentioned in the last section, the refinement of affinities is fast. 

\section{Experiments}
% \vspace{-3mm}
\subsection{Experimental Setup}\label{sec:exp_setup}
\noindent\textbf{Datasets.} Densely annotating video frames requires intensive manual labeling efforts. The widely used datasets for VSS are Cityscapes~\cite{cordts2016cityscapes} and CamVid~\cite{brostow2009semantic} datasets. However, these datasets only contain sparse annotations, which limits the exploration of temporal information. Fortunately, the Video Scene Parsing in the Wild (VSPW) dataset~\cite{miao2021vspw} is proposed to facilitate the progress of this field. It is currently the largest-scale VSS dataset with 198,244 training frames, 24,502 validation frames and 28,887 test frames. For each video, 15 frames per second are densely annotated for 124 categories. These aspects make VSPW the best benchmark for VSS up till now. Hence, most of our experiments are conducted on VSPW. To further demonstrate the effectiveness of MRCFA, we also show results on Cityscapes, for which only one out of 30 frames is annotated. 

\noindent\textbf{Implementation details.} For the encoder, we use the MiT backbones as in Segformer~\cite{xie2021segformer}, which have been pretrained on ImageNet-1K~\cite{russakovsky2015imagenet}. For VSPW dataset, three reference frames are used, which are 9, 6 and 3 frames ahead of the target, following~\cite{miao2021vspw}. Three-scale features from the last three transformer blocks are used to compute the cross-frame affinities and mine their correlations. For the Mask-based Token Selection (MTS), we set $p$=80\% for MiT-B0 and $p$=50\% for other backbones unless otherwise specified. For training augmentations, we use random resizing, horizontal flipping, and photometric distortion to process the original images. Then, the images are randomly cropped to the size of $480 \times 480$ to train the network. 
% For Cityscapes~\cite{cordts2016cityscapes}, same augmentations are adopted, but the images are randomly cropped to $512 \times 1024$.
We set the batch size as 8 during training. The models are all trained with AdamW optimizer for a maximum of 160k iterations and ``poly'' learning rate schedule. The initial learning rate is 6e-5. For simplicity, we perform the single-scale test on the whole image, rather than the sliding window test or multi-scale test. The input images are resized to $480 \times 853$ for VSPW. We also do not perform any post-processing such as CRF~\cite{krahenbuhl2011efficient}. For Cityscape, the input image is cropped to $512 \times 1024$ during training and resized to the same resolution during inference. And we use two reference frames and four-scale features. The number of frames being processed per second (FPS) is computed in a single Quadro RTX 6000 GPU (24G memory).
\begin{table}[!t]
\centering
\resizebox{0.8\columnwidth}{!}{%
\begin{tabular}{c|c|c|c|c|c|c|c}
% \hline
% \hline\thickhline
 \bottomrule[0.15em]
 \rowcolor{tableHeadGray}
~~~~~Methods~~~~~   & ~~$T$~~ & ~~$t_1$~~ & ~~$t_2$~~ & ~~$t_3$~~ &  ~~mIoU $\uparrow$ ~~ & ~~mVC$_8$ $\uparrow$ ~~ & ~~mVC$_{16}$ $\uparrow$ ~~ \\ \hline \hline
SegFormer~\cite{xie2021segformer}       &   -   & -  & -  & -  &   36.5    &   84.7 &   79.9  \\ \hline \hline
\multirow{7}{*}{MRCFA (Ours)} &2 & -1  & -  &  - & 38.0  & 85.9 &  81.2  \\ \cline{2-3} \cline{4-8} 
   &2 & -3  & -  & - &   38.1   & 85.5  &    80.7   \\ \cline{2-3} \cline{4-8} 
   & 2& -6  & -  & - &   38.2   &    85.1  & 80.3      \\ \cline{2-3} \cline{4-8} 
   & 2& -9  & -  & - & 37.4     & 85.5    &    81.2    \\ \cline{2-3} \cline{4-8} 
   & 3& -6  &  -3  &  -  & 38.4  & 87.0 & 82.1   \\ \cline{2-3} \cline{4-8} 
   & 3 & -9  & -6  & -  & 38.4  & 86.9  & 82.0        \\ \cline{2-3} \cline{4-8} 
   &4 & -9  & -6  & -3  &     \textbf{38.9}  & \textbf{88.8} &  \textbf{84.4}          \\ \hline
\end{tabular}}
\caption{The impact of the selection of reference frames.}
\label{table:albation_reference} 
% \vspace{-10mm}
\end{table}
\begin{table}[!t]
\centering
\resizebox{0.8\columnwidth}{!}{%
\begin{tabular}{c|c|c|c|c|c}
% \hline
\bottomrule[0.15em]
 \rowcolor{tableHeadGray}
~~~$p$~~~     & ~mIoU $\uparrow$ ~ & ~mVC$_8$ $\uparrow$~ & ~mVC$_{16}$ $\uparrow$ ~ & ~Memory (M) $\downarrow$ ~ & ~FPS (f/s) $\uparrow$ ~ \\ \hline \hline
100\% & 39.4 & 89.2 &  84.9     &  1068   &   32.9  \\ \hline
90\%  & 39.1     &  89.1    &  84.8     &  1035   &  34.2   \\ \hline
70\%  & 39.1     & 88.2     & 83.9      &  969   &   36.8  \\ \hline
\textcolor{red}{50\%}  & \textcolor{red}{38.9}  & \textcolor{red}{88.8}     &  \textcolor{red}{84.4}     &  \textcolor{red}{903} \color{red}{(15.4\%)}  &  \textcolor{red}{40.1} \color{red}{(21.9\%)} \\ \hline
30\%  & 38.5     & 86.7     & 81.9      &  838  &   43.5  \\ \hline
10\%  & 35.9     & 86.2     & 81.7      &  773   &   47.2  \\ \hline
\end{tabular}}
\caption{The impact of token selection ratio $p$.
% GPU memory is computed using the input size of inference period, \ie, $480 \times 853$. 
The row which best deals with the trade-off between performance and computation resources is shown in \textcolor{red}{red}. }
% \vspace{-8mm}
\label{table:albation_ratio_p} 
\end{table}
% \noindent\textbf{Implementation details.} Our implementation is based on the \textit{mmsegmentation}~\cite{mmseg2020} repository. \textit{Encoder:} we use the MiT backbones as in Segformer~\cite{xie2021segformer}, which have been pretrained on ImageNet-1K dataset~\cite{russakovsky2015imagenet}. \\

\noindent\textbf{Evaluation metrics.} To evaluate the segmentation results, we adopt the commonly used metrics of Mean IoU (mIoU) and Weighted IoU (WIoU), following~\cite{shelhamer2017fully}. We also use Video Consistency (VC)~\cite{miao2021vspw} to evaluate the category consistency among the adjacent frames in the video, following \cite{miao2021vspw}. Formally, video consistency VC$_n$ for $n$ consecutive frames for a video clips $\{\bm{I}_c\}_{c=1}^{C}$, is computed by: $\text{VC}_{n}=\frac{1}{C-n+1}\sum_{i=1}^{C-n+1}\frac{(\cap_{i}^{i+n-1}\bm{S}_i)\cap (\cap_{i}^{i+n-1}\bm{S}^{'}_i)}{\cap_{i}^{i+n-1}\bm{S}_i}$,
% \begin{align}\label{Eq:comp_vc}
% \begin{split}
%         \text{VC}_{n}=\frac{1}{C-n+1}\sum_{i=1}^{C-n+1}\frac{(\cap_{i}^{i+n-1}\bm{S}_i)\cap (\cap_{i}^{i+n-1}\bm{S}^{'}_i)}{\cap_{i}^{i+n-1}\bm{S}_i},
% \end{split}
% \end{align}
where $C\geq n$. $\bm{S}_i$ and $\bm{S}^{'}_{i}$ are the ground-truth mask and predicted mask for $i^{th}$ frame, respectively. We compute the mean of video consistency VC$_n$ for all videos in the dataset as mVC$_n$. Following \cite{miao2021vspw}, we compute mVC$_8$ and mVC$_{16}$ to evaluate the visual consistency of the predicted masks. Please refer to \cite{miao2021vspw} for more details about VC.\\
% with ground truth masks $\{\bm{S}_c\}_{c=1}^{C}$ and predicted masks $\{\bm{S}^{'}_{c}\}_{c=1}^{C}$

% \vspace{-8mm}
\subsection{Ablation Studies}
We conduct ablation studies on the large-scale VSPW dataset~\cite{miao2021vspw} to validate the key designs of MRCFA. For fairness, we adopt the same settings as in \secref{sec:exp_setup} unless otherwise specified. The ablation studies are conducted on MiT-B1 backbone.

\noindent\textbf{Influence of the reference frames.} We study the performance of our method with respect to different choices of reference frames in \tabref{table:albation_reference}. We have the following observations. First, using a single reference frame largely improves the segmentation performance (mIoU). For example, when using a single reference frame which is 3 frames ahead of the target one, the mIoU improvement over the baseline (SegFormer) is 1.6\%, \ie, 38.1 over 36.5. Further adding more reference frames, better segmentation performance is observed. The best mIoU of 38.9 is obtained when using reference frames of 9, 6, and 3 frames ahead of the target. Second, for the prediction consistency metrics (mVC$_8$ and mVC$_{16}$), the advantage of exploiting more reference frames is more obvious. For example, using one reference frame ($t_1=-6$) gives mVC$_8$ and mVC$_{16}$ of 85.1 and 80.3, improving the baseline by 0.4\% and 0.4\%, respectively. However, when using three reference frames ($t_1=-9$, $t_2=-6$, $t_3=-3$), the achieved mVC$_8$ and mVC$_{16}$ are much more superior to the baseline, improving by 4.1\% and 4.5\%. The results are reasonable because using more reference frames gives the model a bigger view of the previously predicted features and thus generates more consistent predictions. \\
% we observe that significant improvement is obtained when exploiting more reference frames.\\
\noindent\textbf{Influence of token selection ratio $p$.} We study the influence of the token selection ratio $p$ in terms of performance and computational resources in \tabref{table:albation_ratio_p}. Smaller $p$ represents that less number of tokens in the key features are selected and thus less computation resource is required. Hence, there is a trade-off between the segmentation performance and the required resources (GPU memory and additional latency). In the experiments, when reducing $p=100\%$ to $50\%$, the performance reduces slightly (0.5 in mIoU) while the GPU memory reduces by 15.4\% and FPS increases by 21.9\%. When further reducing $p$ to $10\%$, the performance largely decreases in terms of mIoU, mVC$_8$ and mVC$_{16}$. The reason is that too many tokens are discarded in the reference frames and the remained tokens are not informative enough to provide the required contexts for segmenting the target frame. To sum up, the best trade-off is achieved when $p=50\%$.

\noindent\textbf{Influence of the feature scales.} For VSPW dataset, we use three-scale features output from the last three transformer blocks. Here, we conduct an ablation study on the impact of the used feature scales. The results are shown in \tabref{table:albation_L}. It can be observed that using the features from the last stage ($L=1$) or the last two stages ($L=2$) gives inferior performance while consuming less computational resources and achieving faster running speed. When using three-scale features, the best results are achieved in terms of mIoU, mVC$_{8}$, and mVC$_{16}$. This is due to the fact that the features in different scales contain complementary information, and the proposed affinity decoder successfully mines this information through learning correlations between multi-scale affinities.\\

\begin{table}[!t]
\centering
\resizebox{0.7\columnwidth}{!}{%
\begin{tabular}{c|c|c|c|c|c}
% \hline
\bottomrule[0.15em]
 \rowcolor{tableHeadGray}
~~$L$~~ & ~mIoU $\uparrow$~ & ~mVC$_{8}$ $\uparrow$~ & ~mVC$_{16}$ $\uparrow$~ & ~Params (M) $\downarrow$~ & ~FPS (f/s) $\uparrow$~ \\ \hline \hline
1 & 37.5  &     87.7 & 83.1      &  14.8     &  44.3  \\ \hline
2 & 38.1     & 87.5     & 82.5      &  15.3   & 43.8    \\ \hline
3 & \textbf{38.9} & \textbf{88.8}  & \textbf{84.4}  &   16.2    & 40.1  \\ \hline
\end{tabular}}
\caption{Ablation study on the number of feature scales ($L$). Using more scales of features for our method progressively increases the performance.}
% \vspace{-8mm}
\label{table:albation_L} 
\end{table}
\begin{table}[!t]
\centering
\resizebox{0.8\linewidth}{!}{%
\begin{tabular}{c|c|c|c|c|c|c}
% \hline
\bottomrule[0.15em]
 \rowcolor{tableHeadGray}
Methods                           & ~SAR~ & ~MAA~ & ~mIoU $\uparrow$~ & ~mVC$_{8}$ $\uparrow$~ & ~mVC$_{16}$ $\uparrow$~ & ~Params (M) $\downarrow$~ \\ \hline \hline
SegFormer                         & -   & -   & 36.5     & 84.7     & 79.9     &  13.8   \\ \hline
Feature Pyramid        & -   & -   &   37.8   & 87.0     &  82.0     &  16.2    \\ \hline \hline
\multirow{3}{*}{Affinity Decoder} &  \cmark   & \xmark    &  37.8  &    87.1  & 82.6  &  16.2   \\ \cline{2-7} 
                                  & \xmark    & \cmark    & 37.4  & 88.3 &  83.6 &  16.2   \\ \cline{2-7} 
                                  &  \cmark   & \cmark  &  \textbf{38.9}  &  \textbf{88.8}    &  \textbf{84.4}     &   16.2   \\ \hline
\end{tabular}}
\caption{Ablation study on the affinity decoder. 
% The proposed affinity decoder clearly outperforms the strong baseline of Feature Pyramid. 
Within our design, SAR and MAA are essential parts which contribute to the refinement of the affinity.}
% \vspace{-10mm}
\label{table:albation_affinity_decoder} 
\end{table}

% \begin{table}[!t]
\noindent\textbf{Ablation study on affinity decoder.} We conduct ablation studies on the proposed affinity decoder. The results are shown in \tabref{table:albation_affinity_decoder}. Our affinity decoder processes the multi-scale affinities and generates a refined affinity matrix for each pair of the target and reference frames. It is reasonable to ask whether this design is better than the feature pyramid baseline. For this baseline (Feature Pyramid), we first compute the features for the target frame using the reference frame features at each scale and then merge those multi-scale features. For fair comparisons, we use a similar number of parameters for this baseline and other settings are also the same as ours. The result shows that while Feature Pyramid performs favorably over the single-frame baseline, our approach clearly surpasses it. It validates the effectiveness of the proposed affinity decoder. 

As presented in \secref{sec:method_sar}, our affinity decoder has two modules: Single-scale Affinity Refinement (SAR) and Multi-scale Affinity Aggregation (MAA). The ablation study of two modules is provided in \tabref{table:albation_affinity_decoder}. Only using SAR, our method obtains the mIoU of 37.8, while only using MAA gives the mIoU of 37.4. Both variants are clearly better than the baseline, validating their effectiveness. Combining both modules, the proposed approach achieves the best mIoU, mVC$_{8}$, and mVC$_{16}$. It shows that both SAR and MAA are essential parts of the affinity decoder to learn better affinities to help segment the target frame. 
\begin{table*}[!t]
\centering
\resizebox{0.98\linewidth}{!}{%
\begin{tabular}{c|c|c|c|c|c|c|c}
% \hline
\bottomrule[0.15em]
 \rowcolor{tableHeadGray}
Methods     & Backbone                   & ~~mIoU $\uparrow$~~     & Weighted IoU  $\uparrow$        & mVC$_8$  $\uparrow$    & mVC$_{16}$      $\uparrow$       & Params (M)  $\downarrow$          & ~FPS (f/s) $\uparrow$  ~  \\ \hline \hline
                                % \rotatebox[origin=c]{90}
 SegFormer~\cite{xie2021segformer}  & MiT-B0       & 32.9       & 56.8     & 82.7      & 77.3   &  3.8   &  73.4 \\ \cline{1-8} 
 SegFormer~\cite{xie2021segformer}  & MiT-B1           & {36.5} & 58.8                  & {84.7}     & {79.9}        &  13.8   &  58.7
    \\ \cline{1-8} 
 MRCFA (Ours) & MiT-B0  & 35.2 & 57.9 & 88.0  & 83.2 & 5.2 & 50.0   \\ \cline{1-8} 
 MRCFA (Ours) & MiT-B1 &  \textbf{38.9}  &  \textbf{60.0}   & \textbf{88.8}    &  \textbf{84.4}    & 16.2  &   40.1    \\ \hline \hline
 DeepLabv3+~\cite{chen2017rethinking} & ResNet-101                  & 34.7                      & 58.8                                & 83.2                      & 78.2           &      62.7               &   -   \\ \cline{1-8} 
UperNet~\cite{xiao2018unified}    & ResNet-101            & 36.5                      & 58.6                                   & 82.6                      & 76.1            &    83.2               &     - \\ \cline{1-8} 
PSPNet~\cite{zhao2017pyramid}     & ResNet-101            & 36.5                      & 58.1                                 & 84.2                      & 79.6                &       70.5             &   13.9   \\ \cline{1-8} 
OCRNet~\cite{yuan2020object}     & ResNet-101           & 36.7                      & 59.2                                 & 84.0                      & 79.0                  &     58.1           &   14.3   \\ \cline{1-8} 
ETC~\cite{liu2020efficient}        & PSPNet       & 36.6                      & 58.3                               & 84.1                      & 79.2                     &    89.4            &   -   \\ \cline{1-8} 
NetWarp~\cite{xiao2018unified}    & PSPNet        & 37.0                      & 57.9                                & 84.4                      & 79.4                        &   89.4          &   -   \\ \cline{1-8} 
ETC~\cite{liu2020efficient}        & OCRNet        & 37.5                      & 59.1                                & 84.1                      & 79.1              &  58.1      &   -   \\ \cline{1-8} 
NetWarp~\cite{xiao2018unified}    & OCRNet            & 37.5                      & 58.9                                 & 84.0                      & 79.0        &   58.1   &   -   \\ \cline{1-8} 
TCB$_\text{st-ppm}$~\cite{miao2021vspw}    &   ResNet-101   &      37.5       &     58.6        &     87.0    &  82.1     & 70.5    &   10.0   \\ \cline{1-8} 
TCB$_\text{st-ocr}$~\cite{miao2021vspw}    &  ResNet-101       &     37.4       &     59.3      &   86.9      &  82.0     &   58.1  &   5.5  \\ \cline{1-8} 
TCB$_\text{st-ocr-mem}$~\cite{miao2021vspw}    &  ResNet-101     &   37.8    &   59.5       &      87.9          &     84.0      &   58.1   &   5.5     \\ \cline{1-8} 
SegFormer~\cite{xie2021segformer}  & MiT-B2 &   43.9   & 63.7 &   86.0  &   81.2    & 24.8  &   39.2   \\ \cline{1-8} 
SegFormer~\cite{xie2021segformer}  & MiT-B5  &  48.2    & 65.1 &   87.8  &   83.7  & 82.1   &  17.2  \\ \cline{1-8} 
%  0.4881, 0.6582, 0.8810, 0.8416
% & SegFormer~\cite{xie2021segformer}  &  &  &      &  &     &      &      \\ \cline{2-9} 
MRCFA (Ours) & MiT-B2 & 45.3 &  64.7    & 90.3 &    86.2  & 27.3  &  32.1    \\ \hline
MRCFA (Ours) & MiT-B5 & \textbf{49.9}  & \textbf{66.0} &  \textbf{90.9} &  \textbf{87.4}  &  84.5  & 15.7  \\ \hline
% 0.4998, 0.6595,0.9117, 0.8773
\end{tabular}}
\caption{State-of-the-art comparison on the VSPW~\cite{miao2021vspw} validation set. MRCFA outperforms the compared methods on both accuracy (mIoU) and prediction consistency.}
\label{table:results_sota}
% \vspace{-8mm}
\end{table*}

% \begin{table}[!t]
% \centering
% \resizebox{0.89\linewidth}{!}{%
% \begin{tabular}{c|c|c|c|c|c|c|c}
% % \hline
% \bottomrule[0.15em]
%  \rowcolor{tableHeadGray}
% Methods                           & ~SAR~ & ~CAM~ & ~mIoU $\uparrow$~ & ~mVC$_{8}$ $\uparrow$~ & ~mVC$_{16}$ $\uparrow$~ & ~Params (M) $\downarrow$~ & ~FPS $\uparrow$~ \\ \hline \hline
% SegFormer                         & -   & -   & 36.5     & 84.7     & 79.9     &  13.8    &     \\ \hline
% Feature Pyramid        & -   & -   &   37.7   & 87.6     &  82.8     &  16.4   &     \\ \hline \hline
% \multirow{3}{*}{Affinity Decoder} &  \cmark   & \xmark    &  37.8  &    87.1  & 82.6  &   16.2  &  44.3   \\ \cline{2-8} 
%                                   & \xmark    & \cmark    & 37.4  & 88.3 &  83.6 &  16.2   &  45.1  \\ \cline{2-8} 
%                                   &  \cmark   & \cmark  &  \textbf{38.9}  &  \textbf{88.8}    &  \textbf{84.4}     &   16.2    & 40.1 \\ \hline
% \end{tabular}}
% \caption{Ablation study on the affinity decoder.}
% \label{table:albation_affinity_decoder} 
% \end{table}

% \vspace{-6mm}
\subsection{Segmentation Results}
The state-of-the-art comparisons on VSPW~\cite{miao2021vspw} dataset are shown in \tabref{table:results_sota}. Besides segmentation performance and visual consistency of the predicted masks, we also report the model complexity and FPS. According to the model size, the methods are divided into two groups: small models and large models.
\begin{wraptable}{r}{6.8cm}
\centering
% \vspace{-4mm}
\resizebox{1.0\linewidth}{!}{%
\begin{tabular}{c|c|c|c|c}
% \hline
\bottomrule[0.15em]
 \rowcolor{tableHeadGray}
Methods & Backbone   & mIoU $\uparrow$  & Params (M) $\downarrow$  & FPS (f/s) $\uparrow$  \\ \hline \hline
FCN~\cite{shelhamer2017fully}      & MobileNetV2      & 61.5 &    9.8    & 14.2 \\ \hline
CC~\cite{shelhamer2016clockwork}      & VGG-16     & 67.7   &    -   & 16.5 \\ \hline
DFF~\cite{zhu2017deep}     & ResNet-101   & 68.7 &   -    & 9.7  \\ \hline
GRFP~\cite{nilsson2018semantic}     & ResNet-101   & 69.4  &     -  & 3.2  \\ \hline
% PSPNet18~\cite{zhao2017pyramid}    & ResNet-18 &  13.2   & 69.8 &  -    &   -  &  9.5  \\ \hline
PSPNet~\cite{zhao2017pyramid}    & MobileNetV2   & 70.2 &  13.7   &  11.2  \\ \hline
DVSN~\cite{xu2018dynamic}     & ResNet-101    & 70.3  &  -   & 19.8 \\ \hline
Accel~\cite{jain2019accel}    & ResNet-101    & 72.1  &  -   & 3.6  \\ \hline
ETC~\cite{liu2020efficient}    & ResNet-18    & 71.1  &  13.2   & 9.5  \\ \hline \hline
SegFormer~\cite{xie2021segformer}    & MiT-B0  &   71.9 &  3.7  & 58.5  \\ \hline 
MRCFA (Ours) & MiT-B0 &  72.8  & 4.2 & 33.3 \\ \hline \hline
SegFormer~\cite{xie2021segformer}    & MiT-B1   &   74.1 & 13.8  &  46.8 \\ \hline 
MRCFA (Ours) & MiT-B1 &  \textbf{75.1}  & 14.9 & 21.5 \\ \hline
% CFFM (Ours)    & MiT-B1 &            &  &       &        &   \\ \hline
\end{tabular}}
% \vspace{-3mm}
\caption{State-of-the-art comparison on the Cityscapes~\cite{cordts2016cityscapes} val set.}
\label{table:results_city}
% \vspace{-6mm}
% \end{table}
\end{wraptable}%
Among all methods, our MRCFA achieves state-of-the-art performance and produces the most consistent segmentation masks across video frames. For small models, our method on MiT-B1 clearly outperforms the strong baseline SegFormer~\cite{xie2021segformer} by 2.4\% in mIoU and 1.2\% in weighted IoU. In terms of the visual consistency in the predicted masks, our approach is superior to other methods, surpassing the second best method with 4.1\% and 4.5\% in mVC$_{8}$ and mVC$_{16}$, respectively. 
% For large models, MRCFA obtains a mIoU of 50.3, which is better than the second best by 1.5\%. In terms of mVC$_{8}$ and mVC$_{16}$, MRCFA has an advantage of 2.8\% and 3.3\% over its best counterpart. 
For large models, MRCFA shows similar behavior.
The results indicate that our method is effective in mining the relations between the target and reference frames through the designed modules: SAR and MAA. 
% The improved affinities between the target frame and the reference frame bring more informative contexts to segment the target.

Despite that our approach achieves impressive performance, it adds limited model complexity and latency. Specifically, compared to SegFormer (MiT-B2), MRCFA slightly increases the number of parameters from 24.8M to 27.3M and reduces the FPS from 39.2 to 32.1. 
% On MiT-B5, MRCFA has 84.5M parameters and runs at 15.7 f/s while the corresponding baseline has 82.1M parameters with an inference speed of 17.2 f/s. 
The efficiency of our method benefits from the proposed STM mechanism for which we abandon unimportant tokens.

\begin{figure*}[!t]
    \centering
    \includegraphics[width=0.95\linewidth]{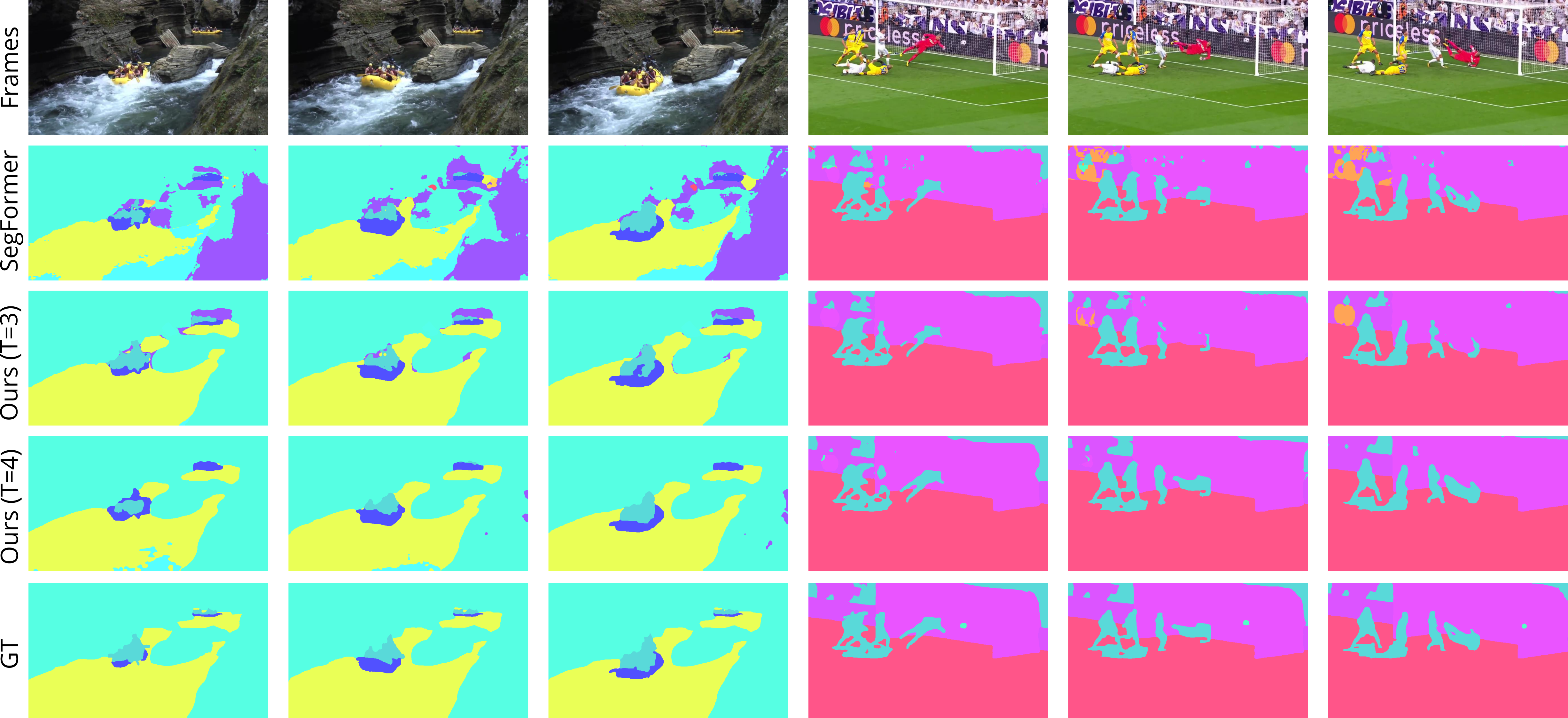}
    % \vspace{-7mm}
    \caption{Qualitative results. From \textit{top} to \textit{bottom}: the input frames, the predicted masks of SegFormer~\cite{xie2021segformer}, the predictions of ours ($T=3, t_1=-3$, $t_2=-6$), the predictions of ours ($T=4, t_1=-3$, $t_2=-6$, $t_3=-9$) and the ground-truth masks. Our model generates better results than the baseline in terms of accuracy and VC. 
    % The proposed method also benefits from more reference frames.
    }
    \label{fig:qual}
    % \vspace{-6mm}
\end{figure*}

We conduct additional experiments on the semi-supervised Cityscapes~\cite{cordts2016cityscapes} dataset, for which only one frame in each video clip is pixel-wise annotated. \tabref{table:results_city} shows the results. Similar to VSPW, MRCFA also achieves state-of-the-art results among the compared approaches under the semi-supervised setting and has a fast running speed. Besides the quantitative comparisons analyzed above, we also qualitatively compare the proposed method with the baseline on the sampled video clips in \figref{fig:qual}. For the two samples, our method generates more accurate segmentation masks, which are also more visually consistent. 
% The results further validate the effectiveness of the proposed designs in mining the contextual information in the reference frames.

\section{Conclusions}
This paper presents a novel framework MRCFA for VSS. Different from previous methods, we aim at mining the relations among multi-scale Cross-Frame Affinities (CFA) in two aspects: single-scale intrinsic correlations and multi-scale relations. Accordingly, Single-scale Affinity Refinement (SAR) is proposed to independently refine the affinity of each scale, while Multi-scale Affinity Aggregation (MAA) is designed to merge the refined affinities across various scales. To reduce computation and facilitate MAA, Selective Token Masking (STM) is adopted to sample important tokens in keys for the reference frames. Combining all the novelties, MRCFA generates better affinity relations between the target and the reference frames without largely adding computational resources. Extensive experiments demonstrate the effectiveness and efficiency of MRCFA, by setting new state-of-the-arts. The key components are validated to be essential for our method by ablation studies. Overall, our exploration of mining the relations among affinities could provide a new perspective on VSS.

\bibliographystyle{splncs04}
\bibliography{egbib}

\end{document}